# Visual Attention and its Intimate Links to Spatial Cognition

John K. Tsotsos, Iuliia Kotseruba, Amir Rasouli, Markus D. Solbach
Dept. of Electrical Engineering and Computer Science
York University, Toronto Canada

**Abstract**
It is almost universal to regard attention as the facility that permits an agent, human or machine, to give priority processing resources to relevant stimuli while ignoring the irrelevant. The reality of how this might manifest itself throughout all the forms of perceptual and cognitive processes possessed by humans, however, is not as clear. Here we examine this reality with a broad perspective in order to highlight the myriad ways that attentional processes impact both perception and cognition. The paper concludes by showing two real world problems that exhibit sufficient complexity to illustrate the ways in which attention and cognition connect. These then point to new avenues of research that might illuminate the overall cognitive architecture of spatial cognition.

## Introduction

This presentation cuts through several major literatures. Obviously, the experimental literature on visual attention is central and is vast (e.g., Itti et al. 2005; Nobre & Kaster 2014). Theoretical and machine models of attention attempt to formalize experimental findings by presenting mechanistic processes that embody them (e.g., Bylinskii et al. 2016; Borji & Itti 2013). Cognitive Architectures present frameworks as well as functional systems that include attentive capabilities and show the interactions of those capabilities with other cognitive components such as working memory, decision-making and action (e.g., Kotseruba & Tsotsos 2018). Robots that use active vision demonstrate embodied systems that can attend during the course of solving tasks (e.g., Bajcsy et al. 2018). It would not be feasible to present good overviews of each and the reader is encouraged to consider the cited sources for reviews.

Perhaps the best early statement of the overall context we consider was given by Barrow and Popplestone (1971): "...consider the object recognition program in its proper perspective, as a part of an integrated cognitive system. One of the simplest ways that such a system might interact with the environment is simply to shift its viewpoint, to walk round an object. In this way, more information may be gathered and ambiguities resolved. A further, more rewarding operation is to prod the object, thus measuring its range, detecting holes and concavities. Such activities involve planning, inductive generalization, and indeed, most of the capacities required by an intelligent machine." This statement can easily be extended beyond object recognition tasks. MacMillan and Creelman (2005) or Carroll (1993) present significant numbers of perceptual tasks, both experimental and practical, that require such extensions combining perception with cognition. These, in part, led Bajcsy et al. (2018) to present the following definition: *An agent is an active perceiver if it knows why it wishes to sense, then chooses what to perceive, and determines how, when and where to achieve that perception*. But this does not seem to encompass the breadth of cognition because it represents perceptual acts in isolation.

Spatial cognition involves the acquisition, organization, use, and maintenance of knowledge about the spatial dimension of real world environments. These capabilities enable humans to accomplish basic and high-level tasks in everyday life. Examples include determining where we are, how we obtain resources, and how we find our way home, and generally how we sense, interpret, behave in, and communicate about space (Waller & Nadel 2013). It is important that spatial cognition is not restricted to navigation and maps. It is neither so simple nor restricted as building an internal map of space and keeping track of one's position on it.

Spatial cognition involves more than just perception; it involves integrating perception with memory, decision-making and task execution. In more lay terms, it seems to involve figuring things out spatially. But what could it mean to figure things out? *Figure out* is a verb meaning to find an answer for, through reasoning. But what type of reasoning? One prominent form of reasoning is deduction. An example of a deductive argument is: *All men are mortal. Socrates is a man. Therefore, Socrates is mortal*. Deduction does not need online verification and can be employed based solely on contents of agent's knowledge base. The knowledge base can be constructed with previously acquired data. On the other hand, consider inductive reasoning. An example of an inductive argument is: *Some men are named Socrates. Therefore, the next man I meet might be named Socrates*. Inductive reasoning takes specific information (premises) and makes a broader generalization (conclusion) that might be considered probable. An agent thus has 2 choices: act based on belief or act to verify. In the real world, induction needs online verification and this would integrate an agent's knowledge with the context of its current situation. Context is the sum total of all that can be sensed and all this is known about the environment in which the agent is found. The magnitude of all of this information necessitates attentional processes. The time-to-respond constraints plus the resource constraints mean that all information cannot be processed all the time with sufficient rapid processing time. The agent must choose which information and how to best manage resources in order to satisfy its constraints. The target of the inductive inference would point to what needs to be sensed.

This paper cannot address the full breadth of all of the issues raised above. However, what it will attempt to do is to focus on attention and its role, broadly speaking, in spatial cognition. Although, as mentioned, it is universal to regard attention as the facility that permits an agent, human or machine, to give priority processing resources to relevant stimuli while ignoring the irrelevant, there is actually much more to it. In Tsotsos (2011), attention is the process by which the brain controls and tunes information processing. Although this definitely includes selecting the relevant and ignoring the irrelevant, there is far more and the next section will briefly overview this breadth.

## The Breadth of Attentional Processing

There is a very large literature on attention, and here, only some highlights will be mentioned as they relate to the main point we wish to present. Kahneman, a pioneer in the area, was very concerned with the optimality of attentional allocation (Kahneman 1970). Allport (1993) is much closer to our own view, saying it is pointless to focus on the locus of selection and on which processes do and do not require attention. He claims that attentional functions are of very many different kinds, serving a great range of computational functions. Egeth and Yanits (1997) in a classic paper, suggest that the key issues are: 1. control of attention by top-down (or goal-directed) and bottom-up (or stimulus-driven) processes; 2. representational basis for visual selection, including how much attention can be said to be location or object based; 3. time course of attention as it is directed to one stimulus after another. Corbetta and Shuman (2002) consider the brain regions involved and conclude that partially segregated networks carry out different attentional functions. These are: the intraparietal and superior frontal cortex prepare and apply goal-directed selection for stimuli and responses; and, the temporoparietal cortex and inferior frontal cortex specialized for detection of behaviourally relevant stimuli. Rossi et al. (2009) claim that frontal and parietal cortices are involved in generating top-down control signals for attentive switching, which may then be fed back to visual processing areas. The PFC in particular plays a critical role in the ability to switch attentional control on the basis of changing task demands. For Miller and Buschman (2013), visual attention may be focused via a frontoparietal network acting on the visual cortex. These network interactions may be regulated via rhythmic oscillations. The brain may operate discretely with pulses of activity routing packets of information. Oscillations are limited in bandwidth and may explain why conscious thought is limited in capacity. Most of these works and many others do not, however, provide a mechanistic view of attention, in the sense of Brown (2014): "most ask what brain regions are active during attentive processes or what networks are active instead of what mechanisms are necessary to reproduce the essential functions and activity patterns in an attentive system." We seek a mechanistic explanation that could be embodied in a realistic agent to exhibit those same functions, activity patterns, and behaviours.

It seems that we need to be explicit about the attentional functions under consideration. The following are the basic attentional functions, adapted and augmented from Tsotsos (2011)[1]:
**Alerting** The ability to process, identify, and move attention to priority signals.
**Attentional Footprint** Optical metaphors describe the 'footprint' of attentional fixation in image space, and the main ones are Spotlight, Zoom Lens, Gradient, and Suppressive Surround.
**Binding** The process by which visual features are correctly combined to provide a unified representation of an object.
**Covert Attention** Attention to a stimulus in the visual field without eye movements.
**Disengage Attention** The generation of signals that release attention from one focus and prepare for a shift.
**Endogenous Influences** Endogenous influence is an internally generated signal used for directing attention. This includes domain knowledge or task instructions.
**Engage Attention** The actions needed to fixate a stimulus whether covertly or overtly.

---

[1] Pointers to the seminal papers that describe each of these can be found in Tsotsos (2011).

**Executive Control** The system that coordinates the elements into a coherent unit that responds correctly to task and environmental demands including selecting, ordering, initiating, monitoring, and terminating functions.

**Exogenous Influences** Exogenous influence is due to an external stimulus and contributes to control of gaze direction in a reflexive manner. Most common is perhaps the influence of abrupt onsets.

**Inhibition of Return** A bias against returning attention to previously attended location or object.

**Neural Modulation** Attention changes baseline firing rates as well as firing patterns of neurons for attended stimuli.

**Overt Attention** Also known as **Orienting** — the action of orienting the body, head, and eyes to foveate a stimulus in the 3D world. Overt fixation trajectories may be influenced by covert fixations.

**Postattention** The process that creates the representation of an attended item that persists after attention shifts.

**Preattentive Features** The extraction of visual features from stimulus patterns perhaps biased by task demands.

**Priming** Priming is the general process by which task instructions or world knowledge prepares the visual system for input. **Cueing** is an instance of priming; perception is speeded with a correct cue, whether by location, feature, or complete stimulus. Purposefully ignoring has relevance here also and is termed **Negative Priming**. If one ignores a stimulus, processing of that ignored stimulus shortly afterwards is impaired.

**Recognition** The process of interpreting an attended stimulus, facilitated by attention.

**Salience/Conspicuity** The overall contrast of the stimulus at a particular location with respect to its surround.

**Search** The process that scans the candidate stimuli for detection or other tasks among the many possible locations and features in cluttered scenes.

**Selection** The process of choosing one element of the stimulus over the remainder. Selection can be over locations, over features, for objects, over time, and for behavioural responses, or even combinations of these.

**Shift Attention** The actions involved in moving an attentional fixation from its current to its new point of fixation.

**Time Course** The effects of attention take time to appear, and this is observed in the firing rate patterns of neurons and in behavioural experiments, showing delays as well as cyclic patterns.

**Update Fixation History** The process by which the system keeps track of what has been seen and processed which are used in decisions of what to fixate and when.

**Visual Working Memory** Attention seems necessary to select which stimuli and what level of their interpretation are stored in visual working memory. The contents of visual working memory impact subsequent perceptual actions.

In the following, we show two real-world tasks where attentional processes necessarily are prominent with the goal of connecting the above functions to how these tasks might be solved.

## Two Real-World Tasks

To this point, all we have done is present a perspective on the breadth of human visual attention. How exactly these attentional capabilities might be connected to cognition is not a well-studied problem, although it seems widely understood that there must be a connection. In an effort to illuminate these potential connections, we consider two real-world tasks, the first being the common same-different visual discrimination problem but with an active observer in 3D, and the second is the task of how car drivers and pedestrians communicate. The first task is relevant for any kind of robot whose role is to be a real assistant in the home or in manufacturing or medical setting, while the second task is critical for the safe function of autonomous cars. Each also has clear scientific value with the goal of understanding how humans accomplish these tasks.

Carroll (1993) explores the many cognitive abilities of humans and one of those is the same-different task. Given two objects, in our case 3D real objects at arbitrary positions and poses, are they the same or different, how quickly can this determination be made and how many views are needed and from which positions are those views taken? The same-different problem is a standard component of IQ tests or other cognitive ability examinations, so it has been considered as a good indicator of human intelligence. There have been many studies (Carroll looks at several) of human performance, but it seems that in all cases, stimuli are line drawings portrayed in 2D. Certainly, no active perceiver is involved. Not only would extensions to 3D active observers have scientific value, but clearly if we wish an intelligent robot to perform this task we will need to understand such a task's characteristics. Of particular interest for this paper is exactly where does attention come in during its solution? We thus decided to embark on experiments that might illuminate these questions and preliminary results are shown below.

It is important to emphasize that this is a pilot study and thus much further work is needed to confirm any conclusion. Nevertheless, as a pilot study, the experiments revealed several points of value. Using plywood, we constructed a number of objects made out of orthogonal blocks, all painted the same colour (light yellow) and connected only using orthogonal angles. An example is shown in Figure 1a. Object size overall ranges from 19cm x 9cm x 10cm to 19cm x 13cm x 10cm. The 5 different objects we constructed are shown in Figure 1b. Two objects were placed in a light-controlled room (3.2m x 6.1m) with light sources in the room corners as shown in Figure 2. The objects were mounted on a stand (1.4m high) and configured to be 1.1m apart, roughly in the center of the room. Object pose was determined randomly from a restricted set of orientations (25, 45, 90 and 180 degrees along a random axis) and object pairs were randomly generated. 8 subjects participated in the experiment. Subjects were outfitted with a construction helmet with a mounted stereo camera (Stereolabs ZED-M running at 720p@60FPS) as seen in Figure 2. The cameras were calibrated to be 5cm above and 2cm in front of subject's eyes. The recording device was an Alienware Laptop (i7, 16GB, GeForce 1070), carried in a backpack worn by the subject. Subjects were at all times untethered and able to move around completely free. Tracking of the 6DOF of camera pose was achieved by combining the stereo data with data from the built-in IMU in the camera. Subjects were asked to move their head in order to change fixation instead of their eyes with their head fixed (since we cannot observe or measure eye movements if the head is stationary). In this way we were able to use temporal spatial binning of the trajectory to determine if a subject took a view or not.

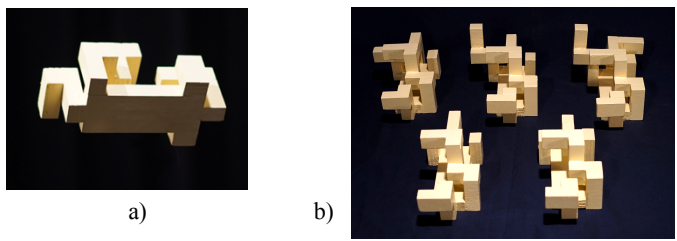

*Figure 1. a) One of the objects in the test; b) the full set of distinct objects used in the experiment.*

a)                b)

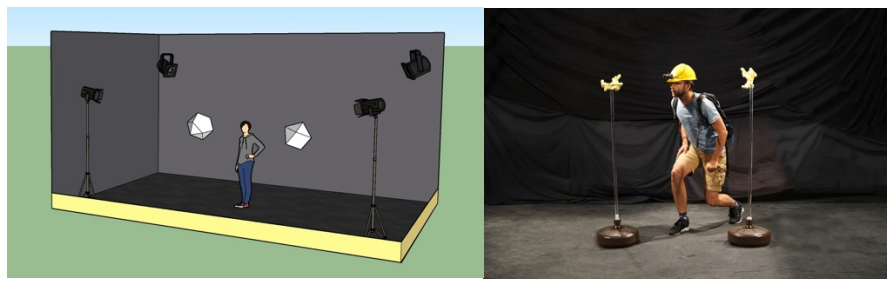

*Figure 2. The experimental setup, the graphic depiction on the left and the actual on the right, showing two objects mounted on two stands, in the test room with the light sources.*

We first wondered if a single 2D view of the scene (i.e., trying to solve the task as a passive observer) with two objects would suffice for the task (Figure 3). The subjects were presented with image pairs (with one object per image), on a paper and were asked to determine whether the objects were the same or different. The results showed that 2D projections of 3D objects are not enough to discriminate the objects. In our experiments with over 60 trials, the subjects performed poorly on this task and achieved an average accuracy of 37.8% while taking around 23.6s for each trial. Self-occlusion makes it hard to perform the same-different task on 3D objects in 2D. The scenarios in which an object is oriented in a way that it self-occludes discriminative parts are difficult to solve based on a single view. Since we forced the subjects to give an answer (speeded 2AFC - same or different), a realistic performance of this task would have been lower if given the possibility of selecting a third option, namely "unable to tell".

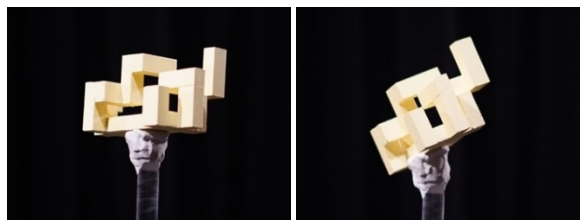

*Figure 3. Samples of the 2D object views used in the passive part of the experiment. Here, we see the same object with a rotational difference of 45 degrees around a random axis.*

However, the 3D, or active observer, version of this experiment gave the subjects the possibility to move freely around the actual 3D objects to observe them from different viewpoints as permitted by the environment (see Figure 2). Figure 4b visualizes a sample run by a subject showing the path followed by the subject around the object highlighted by a dotted line and the viewpoints taken depicted by oriented pyramids (viewing frusta) portraying the direction in which the view is taken. The performance using the same set of objects with same orientations in 3D increased to 80.89% in comparison to the 2D version of the task. Figures 5 and 6 show two examples of the trajectories and viewpoints of three different subjects, superimposed on the top- and front-view of the experimental setup.

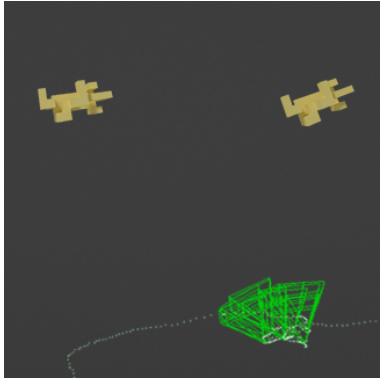
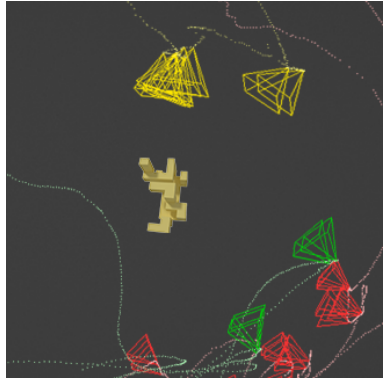

*Figure 4. A visualization of a sample trial. a) The objects are shown at the top of the image and below are the dotted line and pyramids showing the trajectory of the subject's head and viewing frusta respectively. b) The image shows the viewpoints and trajectories of three different subjects while observing the same object. Distinct colours correspond to individual subjects.*

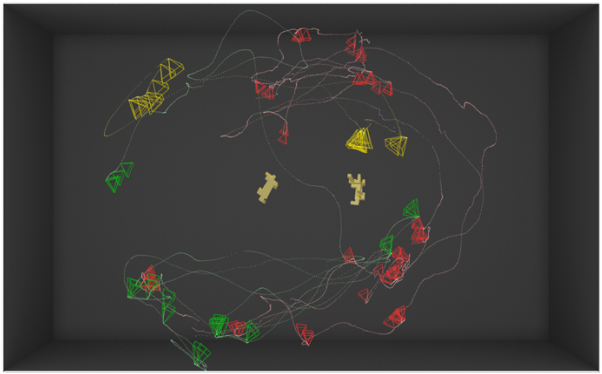
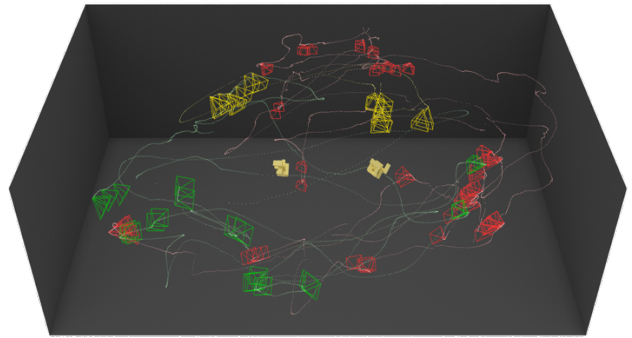

*Figure 5. A visualization of trials performed by three different subjects (red - R, green - G, yellow - Y) under the same conditions. Two different objects were placed in the middle of the room, with poses differing by 90°. The left image shows a top-view rendering of the scene, and the right image shows the front-view of the scene. The trajectory of each subject is plotted in the corresponding colour and each observation of the objects is displayed with an oriented frustum. Subjects R, G and Y took 37, 19, and 15 discriminable views and moved 37.8m, 25.4m, and 10.4m respectively.*

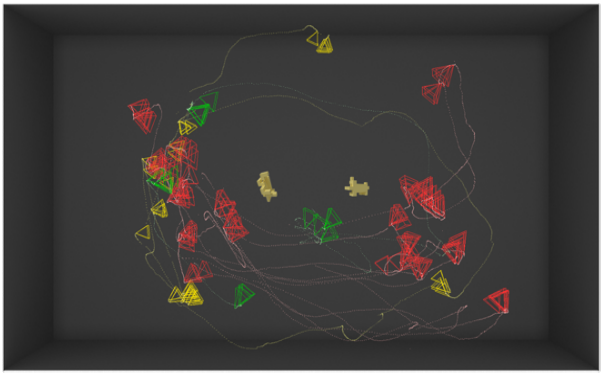
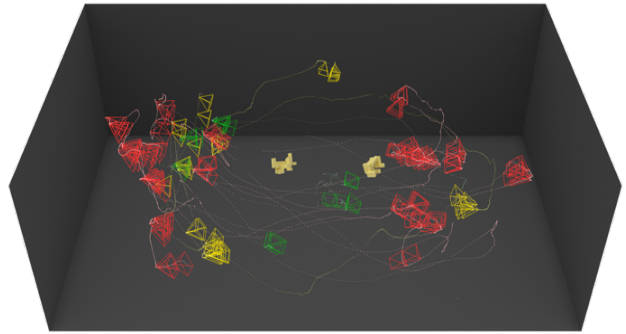

*Figure 6. A visualization of trials performed by three different subjects (red - R, green - G, yellow - Y) under the same conditions. Two different objects were placed in the middle of the room, with poses differing by 180°. The left image shows a top-view rendering of the scene, and the right image shows the from-view of the scene. The trajectory of each subject is plotted in the corresponding colour and each observation of the objects are displayed with an oriented frustum. Subjects R, G and Y took 86, 17, and 26 discriminable views and moved 30.8m, 9.4m, and 15.5m respectively.*

On average, the subjects took 25 views, walked 19.4 m and spent 53.3s to solve one trial. In total over 60 trials were conducted. Figure 7 shows the effect of different orientations on the number of views taken. The results show that increase in the difference of orientations between the objects led to more observations to solve the problem. For the same objects only 25° apart in pose, subjects needed on average 21.6 views. When the pose difference increased to 180°, the subjects on average made more than 29 observations.

The effect of orientation difference on accuracy is presented in Figure 8. The accuracy for 25° difference was about 87%, down to 68.72% as the orientation difference increased to 180°.

Even though these results are preliminary and both more subjects and more analysis are required, we draw the following conclusions:
1. The task requires many views of the scene with no obvious pattern that might lead to some small set of canonical views.
2. What a subject remembers during the trial is clearly important for the integration of successive viewpoints in the course of solving the task, but this experiment does not directly reveal this. Future work will need to address this.
3. The greater the pose difference between objects, the lower the accuracy and the greater the number of views required, whether the objects are the same or different.

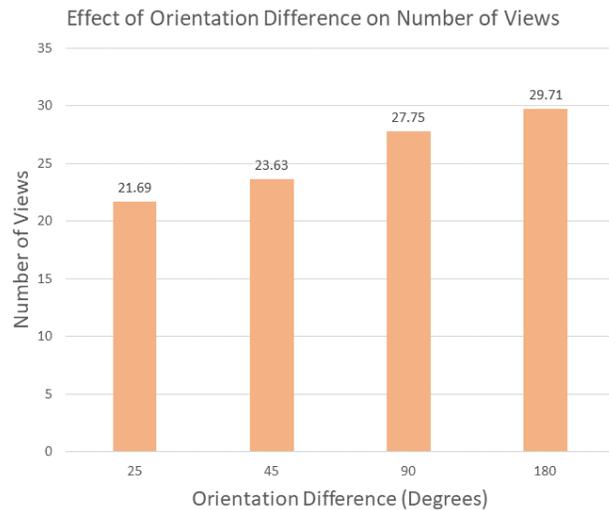

*Figure 7. Increasing orientation difference results in more views taken.*

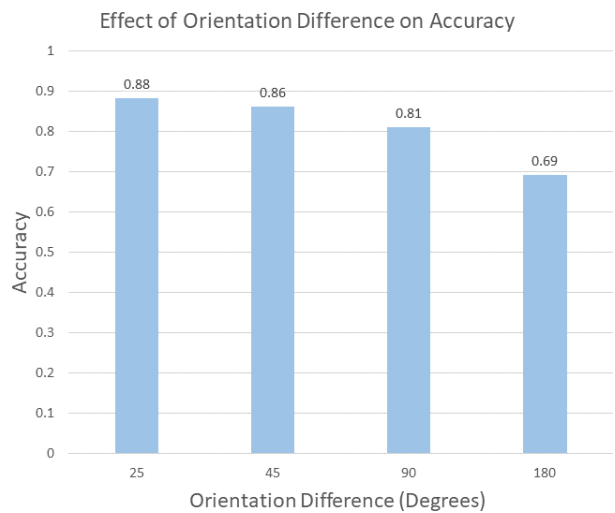

*Figure 8. Increasing orientation difference results in decreasing accuracy.*

We can also suspect that in order to solve the task, a number of the attentional capabilities listed earlier need to be employed. How these are ordered, parameterized and coordinated seems important. Specifically, it seems clear that at least the following, from the list presented earlier, must play a role:

    Disengage Attention - the switch from one view to another necessarily requires a disengagement of attention from the previous view.
    Endogenous Influences - the fact that the task is a speeded 2AFC provides an internal motivation for the actions of the subject.
    Engage Attention - the switch from one view to another requires engagement of attention to the new view.
    Executive Control - during the execution of the task, for 20 - 30 seconds, all of the actions of the subject - walking, viewing, recognition, etc. - obviously require ordering, parameterization, synchronization, monitoring, and more control actions in order to successfully complete the task.
    Exogenous Influences - the task presented to subjects is an instance of an exogenous influence.
    Inhibition of Return - suppressing already viewed images may be useful to not unnecessarily repeat actions. This might be modulated by the way working memory decays.
    Overt Attention (Orienting) - every viewpoint change is an instance of overt attention.
    Priming - subjects might use priming in order to set up the detection of a feature from one object in another in order to match features.
    Recognition - the combination of features seen and their 3D poses into unitary objects is an instance of recognition.
    Search - search is involved in determining the features seen and from which poses to view to process next.
    Selection - new viewpoints must be selected at each position from among the many possible viewpoints.

Shift Attention - attention is shifted from one feature or set of features to another, and from one object to another.
Update Fixation History - what has been seen needs to be tracked to avoid unnecessary views. This, however, is modulated by how working memory is maintained (what is stored, and for how long).
Visual Working Memory - the management of working memory seems critical since viewers take so many views in order to solve the task. Clearly, what is extracted from each view, how it impacts the choice of next views, how what is extracted is integrated into the perception of the whole, how the task-relevant subsets of what enters working memory play a role in decision-making, and more, all connect to several other attentional functions.

This same-different task is a task which humans perform routinely in everyday activities, even though here in this experiment, it is pushed to an extreme. Often, we design objects to be easily discriminable, say by colour or size or pattern, but this is not always the case. For example, consider a task where you are given a part during an assembly task and need to go to a bin of parts in order to find another one of the same. Sometimes this will be easily solved while sometimes it might require picking up candidate objects and checking different viewpoints in order to confirm a match (as Barrow and Popplestone suggested). Certainly, the common LEGO toy requires one to perform such tasks many times while constructing a block configuration, either copying from a plan or mimicking an existing one.

An obvious question in the current scientific environment is whether or not a solution to this problem can be learned. It seems quite likely possible to learn a viewing policy that simply covers all directions of the viewing sphere around each object and then compares feature representations on that sphere. But this obvious brute force solution does not illuminate how humans do it while requiring fewer views. The design of this experiment and its stimuli have as foundation the theoretical work of Kirousis & Papadimitriou (1988) and Parodi et al. (1998) who showed that even to determine whether an image of a polyhedral object is physically realizable has inherent exponential complexity and no learning procedure can change this to make it efficient.

## Driver-Pedestrian Joint Attention

Let us move to a second, even more complex, example. Spatial reasoning is a vital aspect of autonomous driving and benefits a wide range of tasks such as route selection, driving maneuvers, scene understanding and road users' behaviour prediction. Understanding pedestrians' behaviour at crosswalks is particularly important as they are the most vulnerable road users. Among various factors that influence pedestrian crossing decision (Rasouli & Tsotsos 2018) one crucial component is the configuration of the traffic scene in terms of the locations of pedestrians, other road users, the ego-vehicle and traffic signs and signals. More intuitively, when predicting a pedestrian's behaviour, four factors relating to the spatial configuration of the scene should be considered (see Figure 9):

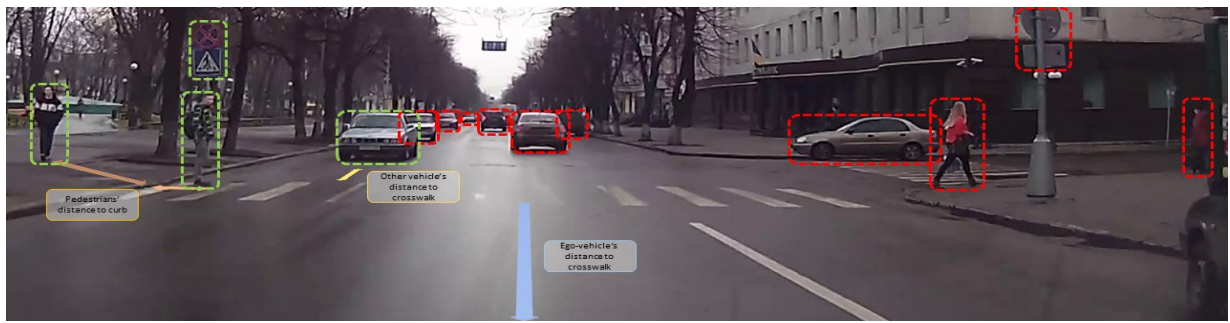

*Figure 9: A typical traffic scenario where the vehicle is driving straight while two pedestrians (on the left) are intending to cross. The elements that are relevant and irrelevant to the task are highlighted with green and red bounding boxes.*

1. The pedestrian's position with respect to the curb (or the crosswalk's boundaries) can indicate whether and where (in the case of intersections) they intend to cross. By standing after the curb, pedestrians often explicitly communicate their intention of crossing to approaching traffic (Rasouli et al. 2018). In addition, being close (or belonging) to a larger group of pedestrians may influence the way the pedestrian will behave. For instance, pedestrians would be more conservative when crossing alone compared to when as a group (Harrell 1991).
2. The positioning of the ego-vehicle is among the most influential factors on pedestrian crossing behaviour (Das et al. 2005). How far the ego-vehicle is from the crosswalk can determine whether the pedestrian intends to cross.

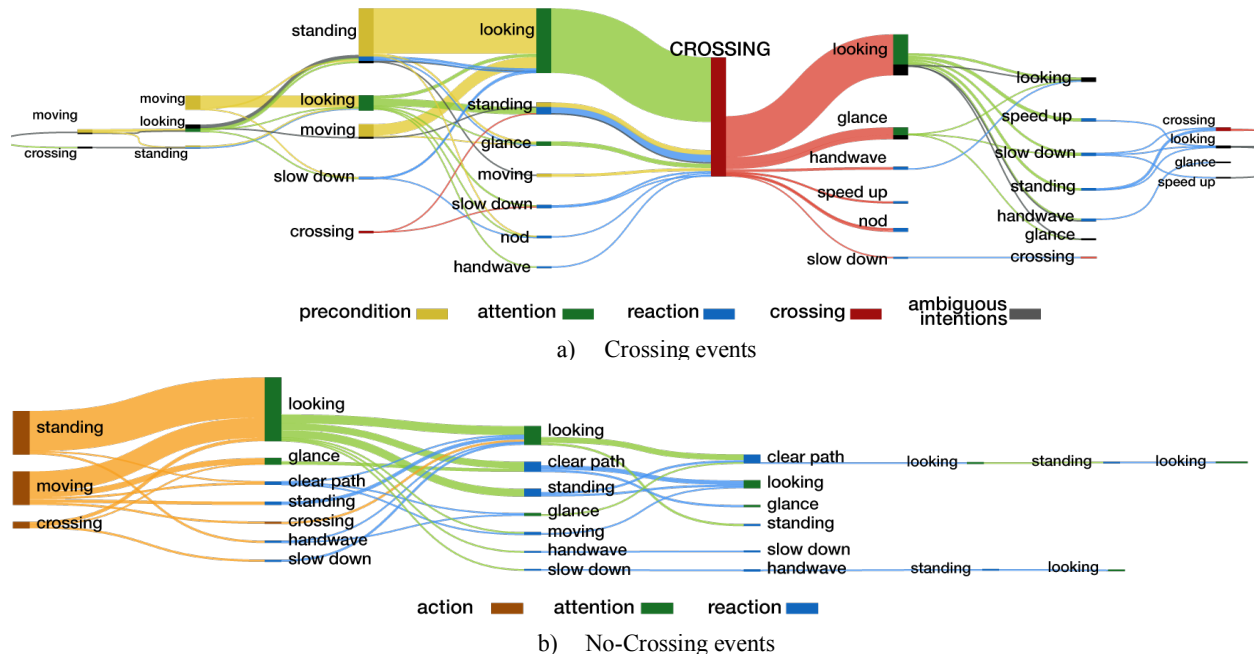

*Figure 10. A visualization of sequences of events observed in the JAAD dataset. Diagram a) shows a summary of 345 sequences of pedestrians' actions before and after crossing takes place. Diagram b) shows 92 sequences of actions which did not result in crossing. Vertical bars represent the start of actions. Different types of actions are colour-coded as the precondition to crossing, attention, reaction to driver's actions, crossing or ambiguous actions. Curved lines between the bars show connections between consecutive actions. The thickness of lines reflects the frequency of the action in the 'crossing' or 'no-crossing' subset. The sequences longer than 10 actions (e.g. when the pedestrian hesitates to cross) are extremely rare, and are truncated from both ends in the figure. (from Rasouli et al. 2018).*

3. On multiple-lane or two-way streets, traffic interactions may involve other vehicles and depending on where they are with respect to the pedestrian (e.g. blocking their way) or the ego-vehicle, would affect the pedestrian's crossing decision.
4. From an algorithmic point of view, it is important to identify which traffic signals or signs relate to the pedestrian's crossing decision. For example, at a four-way intersection, there may be multiple signs and signals at different crosswalks to control the flow of traffic.

As drivers, we seem able to almost immediately and with little effort understand the intentions of a pedestrian and to communicate with them so that the next actions for both car and pedestrian are understood. However, on closer inspection, the visual behaviour of a pedestrian is far from simple or in some sense, standardized. Rasouli et al. (2018) show pedestrian motifs extracted from a large video dataset that portray the complexity and variability of actions for a pedestrian when crossing and when deciding to not cross. Pedestrians might use body movements to communicate, look towards the traffic signaling their intention of crossing or demonstrate different walking patterns depending on the situation, e.g. whether they are crossing at a designated crosswalk or jaywalking. These motifs are reproduced here in Figure 10.

Further, Rasouli et al. (2018), and others, show that the spatial configuration of an intersection is important (this sounds like it should be obvious, but it is satisfying that it is also supported by data). Crossing behaviour depends strongly on the width of the road, whether it is a designated crosswalk or not, whether there are traffic signals, and more. The visual processing of a single glance requires at least 150ms of processing to simply categorize the scene or an object within it. More time is needed for more complex processing (visual search for a complex object in a scene can take minutes) but at a crosswalk one does not have the luxury of a great deal of time and our experience with this task seems to say we can do it in one or a small number of seconds. Clearly, the busier the intersection the more difficult the perceptual tasks of identifying intersection elements. Further, the price of error can be very high so as drivers or pedestrians we are highly motivated to perform the task quickly and correctly.

As in the previous example, we can also suspect that in order to solve this task, a number of the attentional capabilities listed earlier need to be employed. Specifically, it seems clear that at least the following, from the list presented earlier, must play a role in driver-pedestrian interaction:

- Alerting - drivers need to be constantly vigilant for unexpected events so an alerting component is always active.
- Disengage Attention - the switch from one view to another necessarily requires a disengagement of attention from the previous view.
- Endogenous Influences - these are complex for a driver because a driver must balance several goals: to reach a destination; to do so without any kind of incident or accident; to do so without violating traffic rules and conventions; to minimize distraction; to monitor all external movements and actions that might be relevant.
- Engage Attention - a switch from one view to another necessarily requires engagement of attention to the new view.
- Executive Control - during the execution of the driving task, requiring often long periods of time, all of the relevant actions obviously require ordering, parameterization, synchronization, monitoring, and more control actions in order to successfully complete the task.
- Exogenous Influences - drivers are bombarded by external influences whether they be pedestrians, other cars or trucks, cyclists, traffic signals or police, weather, and more, all of which need to be perceived and integrated into task processing.
- Overt Attention - the world external to the driver requires constant monitoring, and human drivers change fixation and gaze frequently (side mirror, rear mirror, forward directions, street signs, pedestrians, etc.)
- Recognition - pedestrians often use different body movements that transmit their intention. Looking at the right place in the scene helps identify these clues to further interpret their behaviour. For instance, pedestrians orient their head towards the road (Figure 9) to indicate their intention of crossing or use hand gestures to communicate, e.g. yield or ask for right of way. Focusing attention to regions where these cues might appear, e.g. upper torso to monitor head and hand movements, can both speed up and enhance the accuracy of recognition.
- Salience/Conspicuity - as illustrated in Figure 9, it is often the case that the majority of the traffic elements in the scene are irrelevant to the task. For instance, in this figure, the vehicle is driving straight, therefore, the pedestrians that are intending to cross from the left side, the first vehicle that is approaching the crosswalk and the crossing sign on the left side of the road are relevant to the pedestrians' crossing decision.
- Search - a driver will search a scene often, perhaps to find a street sign, perhaps looking for a house number, or perhaps to scan for potholes.
- Selection - as multiple agents might be involved in the interaction, it is important to prioritize what to look at first. For instance, in Figure 9, the pedestrian standing on the road and the approaching vehicle should be focused on first, followed by the second pedestrian who is walking towards the curb.
- Priming - a driver's goals (endogenous influences) will prime their decisions and perceptions.
- Shift Attention - attention is shifted from one feature or set of features to another, and from one object to another.
- Update Fixation History - what has been seen needs to be tracked to avoid unnecessary views. This, however, is modulated by how working memory is maintained (what is stored, and for how long).
- Visual Working Memory - the management of working memory seems critical since viewers take so many views in order to solve the task. Clearly, what is extracted from each view, how it impacts the choice of next views, how what is extracted is integrated into the perception of the whole, how the task-relevant subsets of what enters working memory play a role in decision-making, and more, all connect to several other attentional functions.

As argued in the previous example, it is not obvious that a purely data-driven learning strategy will illuminate human cognitive processes. However, a mechanistic model of those processes is still a long way off.

## Conclusions

The main point of this paper is to argue that the definition of attention presented in Tsotsos (2011), namely, *attention is the process by which the brain controls and tunes information processing*, applies much more broadly than just to visual perception. Visual perception is part of active agents, whether biological or machine, and as such must be effectively integrated into an overall cognitive system that behaves purposely in the real word. This points to the need to expand consideration to include active observers and reasoners. It is not difficult to argue against any purely passive perceptual framework. Passive perception gives up control and thus any agent that relies on it cannot assert purpose to its behaviour. A passive sensing strategy, no matter how much data is collected, no matter how high the quality of statistics extracted, gives up control over the specific characteristics of what is sensed and at what time and for which purpose. Passive sensing reduces or eliminates the utility of any form of predictive reasoning strategy (hypothesize-and-test, verification of inductive inferences including Bayesian, etc.). Everyday behaviour relies on sequences of perceptual, decision-making, and physical actions selected from a large set of elemental capabilities. It is known that the problem of selecting a sequence of actions to satisfy a goal under resource constraints is known to be NP-hard (Ye & Tsotsos 2001). Humans are remarkably capable regardless. It is a challenge to discover exactly how they are so capable.

**Acknowledgements**
This research was supported by several sources, via grants to the senior author, for which the authors are grateful: Air Force Office of Scientific Research USA (FA9550-18-1-0054), Office of Naval Research USA (N00178-16-P-0087), the Canada Research Chairs Program (950-231659), and the Natural Sciences and Engineering Research Council of Canada (RGPIN-2016-05352), and the NSERC Canadian Field Robotics Network (NETGP-417354-11).